# Self-Attention Networks for Intent Detection

**Sevinj Yolchuyeva, Géza Németh, Bálint Gyires-Tóth**

Department of Telecommunications and Media Informatics,

Budapest University of Technology and Economics, Budapest, Hungary

{syolchuyeva, nemeth, toth.b}@tmit.bme.hu

## Abstract

Self-attention networks (SAN) have shown promising performance in various Natural Language Processing (NLP) scenarios, especially in machine translation. One of the main points of SANs is the strength of capturing long-range and multi-scale dependencies from the data. In this paper, we present a novel intent detection system which is based on a self-attention network and a Bi-LSTM. Our approach shows improvement by using a transformer model and deep averaging network-based universal sentence encoder compared to previous solutions. We evaluate the system on Snips, Smart Speaker, Smart Lights, and ATIS datasets by different evaluation metrics. The performance of the proposed model is compared with LSTM with the same datasets.

## 1 Introduction and Related Work

Spoken dialogue systems are agents that are intended to help users to access information efficiently by speech interactions (Liu, et al., 2006). In doing so, spoken dialogue systems categorize most of the major fields of spoken language technology, including speech recognition and speech synthesis, language processing, and dialogue system (McTear, 2002). There are different areas of research in the field of spoken dialogue systems. Spoken language understanding (SLU) is one of the essential components of spoken dialogue systems, and it aims to form a semantic frame that captures the semantics of user utterances or queries. Intent detection is one of the main tasks of SLU system. It can be treated as a semantic utterance classification task; since the dialogue system is created to answer a limited range of questions, there is a predefined finite set of intents (Balodis and Deksne, 2019). This task focuses on classifying the user's intent and extracting semantic concepts as constraints for natural language. For example, the utterance "Switch off the garage lights" is related to switching the light off, as shown in Table 1.

Table 1. Example of utterance and a corresponding intent label.

| Utterance | Intent |
|---|---|
| Switch off the garage lights. | SwitchLightOff |
| Get the room brighter, please. | IncreaseBrightness |
| Skip this song and go on to the next one. | NextSong |

Intent detection has been an ongoing field of research in SLU, and similar to most NLP tasks, there are two main approaches to identify the intent of an utterance: rule-based and statistical methods (Hashemi, et al., 2016). The rule-based systems use predefined rules to match new utterances to their intents, and these rules need to be carefully engineered by human experts. Thus, the advancement of these systems requires a huge amount of human effort.
Statistical models, like conditional random field (CRF) and Support Vector Machines, were investigated for this task (Mendoza and Zamora, 2009; Chen et al., 2018). Another important task of SLU is slot filling, which can be formulated as a sequence labelling task. The combination of intent detection and slot filling models was investigated (Mendoza and Zamora, 2009; Kim, 2016).

Furthermore, neural network-based models have also been investigated by (Liu, 2017). Convolutional neural networks (CNN) were applied for classifying intents in (Hashemi, et al., 2016). The combination of CNN and the triangular CRF model (TriCRF) was proposed for the intent labels and the slot filling in (Kim, et al. 2016). During training, the features are learned through CNN layers and shared by the intent detection and slot filling tasks. With this approach, for intent detection, the error on the ATIS dataset was 5.91%, and F1-score was 95.42% for slot filling.

In recent years, neural network-based solutions and word embeddings have gained growing popularity for intent detection (Balodis and Deksne, 2019; Kim, et al. 2016). The enriching word embeddings with semantic lexicons can be helpful intent detection, and it is combined with bidirectional LSTM in (Kim, et al., 2016).

The encoder-decoder neural architectures have achieved remarkable success in various tasks (e.g., speech recognition, text-to-speech synthesis and machine translation). This type of networks has also been enhanced with attention mechanism (Xu, et al., 2015; Luong, et al., 2015). Those models have also been used for intent detection and other SLU tasks (Liu and Lane, 2016; Schumann and Angkititrakul, 2018). The combination of attention-based encoder-decoder architecture and alignment-based methods was studied in (Liu and Lane, 2016) for joint intent detection and slot filling.

Self-attention networks (SANs) have shown outstanding performance in various NLP tasks, such as machine translation (Vaswani et al. 2017), and sentiment analysis (Letarte, et al., 2018) stance classification (Xu, et al., 2018; Raheja and Tetreault 2019). It is a special attention mechanism for selecting specific parts of an input sequence by relating its elements at different positions (Vaswani et al. 2017). With a well-designed architecture, SANs are capable of multi-scale modelling. Inspired by (Xu, et al., 2018), we propose the Self-Attention Network (SAN) architecture for intent detection. In our approach, the self-attention is applied to utterances (input), and it is combined with Bi-LSTM (or LSTM). For evaluation, we used Natural Language Understanding benchmark dataset (Snips) (Goo, et al., 2018), Smart Speaker and Smart Lights dataset (Saade, et al., 2018), and ATIS (Hemphill, et al., 1990). We show the effectiveness of this approach in different experimental settings. The application of pre-trained Word2vec (Mikolov, et al., 2013), and FastText (Bojanowski, et al., 2017) embeddings also helps to get competitive results. The remaining part of the paper is organized as follows. In Section 2, we introduce word embedding methods. Section 3 presents the proposed approach. In Section 4, we describe the datasets, which were used in this work and discuss the experimental setup and the results.

## 2 Word Embedding

Word embeddings map the words to vectors of real numbers. This approach has been widely used as the inputs to neural network-based models for NLP tasks. Word embedding models can be trained with several different tools, such as Word2vec (skip-gram and continuous bag-of-words (CBOW)) (Mikolov, et al., 2013), GloVe (Pennington, et al., 2014), FastText (Bojanowski, et al., 2017). Continuous Bag-of-Words (CBOW) and Continuous Skip-gram models are both powerful techniques for creating word vectors. FastText is one of the recent advances in word embedding algorithms. The main contribution of FastText is to introduce the idea of modular embeddings, which computes a vector for sub-word components, usually n-grams, instead of computing an embedded vector per word. These n-grams are later combined by a simple composition function to compute the final word embeddings. In pre-trained word embedding models, the word embedding tool is trained on large corpora of texts in the given language and highly useful in different NLP tasks. One of the latest embedding methods is Universal Sentence Encoder models (Cer, et al., 2018), which is a form of transfer learning. In (Cer, et al., 2018), it was introduced two encoding models. One of them is based on a Transformer model (TM) and the other one is based on Deep Averaging Network (DAN). They are pre-trained on a large corpus and can be used in a variety of tasks (sentimental analysis, classification, etc.). Both models take a word, sentence or a paragraph as input and generate a 512-dimensional output vector. The transformer-based encoder model targets high accuracy at the cost of greater model complexity and resource consumption (Cer, et al., 2018). But DAN targets performance efficient inference with slightly reduced accuracy.

In this work, we used 300-dimension Word2vec and FastText word embeddings, which were pre-trained on the English Wikipedia corpus. We also investigated TM and DAN based universal encoder models, and each embedding is combined LSTM and Bi-LSTM.

## 3   Proposed Model

In this section, we introduce two models for intent detection:
1. SAN and LSTM (SAN + LSTM)
2. SAN and Bi-LSTM (SAN + Bi-LSTM)

Both proposed models encode each word to its embedding first. We carried out experiments with different embeddings, as discussed in Section 4.3. As the next step, the contextual information in the input sentences (utterances) is encoded. In the first model, LSTM, in the second one, a Bi-LSTM was used to capture the left and right contexts of each word in the input. In the second model, Bi-LSTM combines two unidirectional LSTM layers that process the input from left-to-right and right-to-left, respectively. Both models are followed by the SAN (see Figure 1), which is based on an attention mechanism for selecting specific parts of a sequence by relating its elements at different positions (Vaswani, et al., 2017). In our work, we only perform input-input attention with self-attention. By using the self-attention, the semantics of the entire utterance can be extracted, and it can be helpful for the better classified. To score attention weight vectors we applied the method of (Xu, et al., 2018).

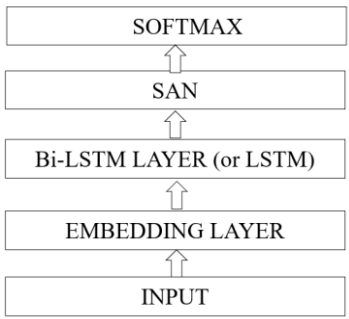

Figure 1. The architecture of the proposed model.

The goal of training is to minimize the loss function. For this purpose, we use multi-class cross-entropy loss,

$$J = -\sum_i \sum_j y_j^i \log \hat{y}_j^i + \gamma \|\theta\|^2 \quad (1)$$

where $i$ is the index of utterance and $j$ is the index of the intent label. $\gamma$ is the $L_2$ regularization coefficient and $\theta$ is the parameter set. $y_j^i$ is the ground-truth label indicator for $i$-th utterance, and $\hat{y}_j^i$ is the predicted probability output of $i$-th utterance. At the output of the network, softmax function was used to predict probabilities.

## 4   Experiments

### 4.1. Dataset

We used Natural Language Understanding benchmark dataset (Snips) (Goo, et al., 2018), Spoken Language Understanding research datasets (Saade, et al., 2018) and Airline Travel Information System (ATIS) dataset (Hemphill, et al., 1990). Snips is a balanced dataset and collected from the Snips personal voice assistant; the number of samples for each intent is approximately the same. The training set contains 13,084 utterances, the test and validation (development set) set consist of 700 - 700 utterances. Vocabulary size is 11,241 and intent types are 7, as shown in Table 2.

Table 2. The intent labels and the number of utterances in each label in Snips.

| Type of intent | Number |
|---|---|
| PlayMusic | 1914 |
| GetWeather | 1896 |
| BookRestaurant | 1881 |
| RateBook | 1876 |
| SearchScreeningEvent | 1851 |
| SearchCreativeWork | 1847 |
| AddToPlaylist | 1818 |

Smart Lights has 6 intents allowing to turn on or off the light or change its brightness or colour, as shown in Table 3. It has a vocabulary size of approximately 400 words. Smart Speaker dataset has 9 intents and vocabulary size is approximately 1,270. The number of utterances in each intent label is presented in Table 4. In these two datasets, we have split the data into 90 % training and 10% test sets. The validation dataset consists of 10% proportion of the training set.

Table 3. The intent labels and the number of utterances in each label in Smart Lights.

| Type of intent | Number |
|---|---|
| Decrease Brightness | 296 |
| Increase Brightness | 296 |
| Set Light Brightness | 296 |
| Set Light Color | 306 |
| Switch Light Off | 299 |
| Switch Light On | 278 |

Table 4. The intent labels and the number of utterances in each label in Smart Speaker.

| Type of intent | Number |
|---|---|
| GetInfos | 199 |
| NextSong | 200 |
| PlayMusic | 1508 |
| PreviousSong | 199 |
| ResumeMusic | 200 |
| SpeakerInterrupt | 172 |
| VolumeDown | 215 |
| VolumeSet | 100 |
| VolumeUp | 260 |

ATIS contains audio recordings of people making flight reservations. The training set contains 4,478 utterances, the test set contains 893 utterances; 500 utterances were used as development set. The intent types in ATIS are unbalanced. For example, the intent atis_flight equals about 73.8 % of the training data, while the number of some intents were less than 10.

### 4.2 Training setup

Data preprocessing may include data normalization, tokenization, lower-casing, removal of punctuation, grammar correction, feature extraction etc., by depending on the task and given dataset. We have done tokenization, have removed punctuation and have converted the numbers to words for all investigated datasets. The word embeddings are initialized with the pretrained 300-dimension Word2Vec or FastText word vectors and these are fixed during training. We also investigated TA and DAN Universal Encoder model-based utterance vectors. The number of units in LSTM and Bi-LSTM is 64.
The L2-regularization coefficient λ in the loss is 0.01.

ADAM (Kingma and Ba, 2015) is used as the optimizer, with a learning rate of 0.001, and with the baseline values of β1, β2 and ε (0.9, 0.999 and 1e-08, respectively). The batch size is 16, the number of epochs is 25.

### 4.3 Evaluation and Results

We evaluated the performance of the models by accuracy, precision, recall, F1-score. The results are presented in Table 5.

By micro and macro averaged, overall F1-scores were computed, and their average was used (Sokolova and Lapalme, 2009). In Table 5, the first column describes the proposed models, and other columns show the overall accuracy and F1-score for each dataset.

For all datasets, SAN + Bi-LSTM consequently have shown better results than SAN + LSTM, as expected. For Snips, the accuracy of FastText + SAN +Bi-LSTM and TM +SAN +Bi-LSTM is almost the same. The result of Word2Vec + SAN + Bi-LSTM, FastText + SAN + Bi-LSTM, and TM + SAN + Bi-LSTM is almost the same for Smart Speaker. The lowest accuracy score for Smart Lights was produced by DAN + SAN + LSTM, which is 90.2%. The highest accuracy score for ATIS was produced by TM + SAN + Bi-LSTM, which is 96.81. This result is comparable with (Goo, et al., 2018; Hakkani-Tür, et al., 2016). We observed that TM based universal encoder can help to improve accuracy.

Furthermore, Figure 2 and Figure 3 show the confusion matrix of Smart Lights and Snips test dataset by using DAN and TM universal encoder vectors with SAN + Bi-LSTM. SAN + Bi-LSTM based DAN correctly classified 30 intents labels out of 31 for SetLightBrightness and 29 tokens out of 30 for SwitchLightOff, which is the same in SAN + Bi-LSTM based TM.

Table 5. Result of proposed models

| Model | Snips | | Smart Lights | | Smart Speaker | | ATIS | |
|---|---|---|---|---|---|---|---|---|
| | Acc(%) | F1-s. | Acc (%) | F1-s. | Acc(%) | F1-s. | Acc(%) | F1-s. |
| Word2Vec + SAN + LSTM | 94.2 | 0.94 | 91.8 | 0.90 | 94.9 | 0.94 | 93.93 | 0.92 |
| FastText + SAN + LSTM | 94.6 | 0.94 | 92.1 | 0.92 | 95.1 | 0.95 | 94.51 | 0.94 |
| DAN + SAN + LSTM | 94.1 | 0.94 | 90.2 | 0.90 | 91.7 | 0.90 | 93.56 | 0.93 |
| TM + SAN + LSTM | 94.2 | 0.94 | 93.6 | 0.93 | 94.2 | 0.93 | 94.81 | 0.94 |
| Word2Vec+SAN+Bi-LSTM | 95.6 | 0.96 | 93.8 | 0.94 | 97.7 | 0.98 | 94.49 | 0.93 |
| FastText + SAN + Bi-LSTM | 96.1 | 0.96 | 93.4 | 0.92 | 97.7 | 0.97 | 95.77 | 0.94 |
| DAN + SAN + Bi-LSTM | 94.2 | 0.94 | 93.2 | 0.93 | 94.7 | 0.95 | 94.91 | 0.93 |
| TM + SAN + Bi-LSTM | 96.5 | 0.97 | 96.6 | 0.97 | 97.7 | 0.98 | 96.81 | 0.95 |

Figure 2. Confusion matrix of Smart Light test dataset by using DAN and TM Universal encoder vectors with SAN + Bi-LSTM.

Figure 3. Confusion matrix of Snips test dataset by using DAN and TM Universal encoder vectors with SAN + Bi-LSTM.

The intent of IncreaseBrightness was predicted correctly in case of 24 out of 30, while 4 intent labels were misclassified to the SwitchLightOn by SAN + Bi-LSTM based DAN.

For Snips, SAN + Bi-LSTM based DAN correctly classified 88 intent labels out of 105 SearchScreeningEvent, while the TM-based approach classified 93 intent labels correctly. The AddToPlaylist, GetWeather, RateBook labels achieved almost the same accuracy from both models. SearchCreativeWork intent labels were better predicted by TM based SAN + Bi-LSTM.

As reasons for the misclassification are that some words can belong to both intent classes, depending on the context, and the size of training data is not large enough.

## Conclusion

In this paper, the combination of SAN and Bi-LSTM for intent detection were proposed. 300-dimensional Word2Vec and FastText embeddings pretrained on English Wikipedia were used as word representations. Utterance vectors of DAN and TM based Universal sentence encoders were investigated. The results were evaluated with the help of accuracy and confusion matrices. Experiments were also carried out with SAN + LSTM, however, the accuracy was worse than with SAN + Bi-LSTM.

Generally, comparison of these models shows that SAN + Bi-LSTM with TM embeddings performs better than other models on all the investigated datasets. In the future, we would like to carry out more comprehensive analysis and investigate other attention mechanisms such as directional self-attention and bi-directional block self-attention (Shen, et al., 2018) for this task.

## Acknowledgements

The research presented in this paper has been supported by János Bolyai Research Scholarship of the Hungarian Academy of Sciences, by Doctoral Research Scholarship of Ministry of Human Resources in the scope of New National Excellence Program, by the BME-Artificial Intelligence FIKP grant of Ministry of Human Resources (BME FIKP-MI/SC) and by the AI4EU project. We gratefully acknowledge the support of NVIDIA Corporation with the donation of the Titan Xp GPU used for this research.